\documentclass[letterpaper, 10 pt, conference]{ieeeconf}  

\IEEEoverridecommandlockouts                              

\pdfoutput=1

\usepackage{graphicx}
\graphicspath{ {fig/} }
\usepackage{subcaption}
\usepackage{amssymb}
\usepackage{mathtools}
\usepackage{caption}
\usepackage{hyperref}
\usepackage{comment}
\usepackage{multirow}
\usepackage{graphicx}
\usepackage{adjustbox}
\usepackage{mathtools}
\usepackage{wrapfig}
\DeclarePairedDelimiterX{\infdivx}[2]{(}{)}{%
  #1\;\delimsize\|\;#2%
}

\usepackage[dvipsnames]{xcolor}
\newcommand{\edits}[1]{\textcolor{black}{#1}}

\begin{document}

\title{\LARGE \bf
It Takes Two: Learning to Plan for Human-Robot Cooperative Carrying
}

\author{Eley Ng$^{1}$, Ziang Liu$^{2}$, and
Monroe Kennedy III$^{1,2}$
\thanks{Authors are members of the ARMLab in the Departments of Mechanical Engineering$^{1}$ and Computer Science$^{2}$, Stanford University, Stanford, CA 94305, USA. 
{\tt\small \{eleyng,ziangliu, monroek\}@stanford.edu.} 
The first author is supported by the NSF Graduate Research Fellowship. \edits{This work was supported by Stanford Institute for Human-Centered Artificial Intelligence (HAI) and conducted under IRB-65022.}  \edits{Links to video and code repositories are on the project webpage: \href{https://sites.google.com/view/cooperative-carrying}{https://sites.google.com/view/cooperative-carrying}.}
}}

\maketitle
\thispagestyle{empty}
\pagestyle{empty}

\begin{abstract}

Cooperative table-carrying is a complex task due to the continuous nature of the action and state-spaces, multimodality of strategies, and the need for instantaneous adaptation to other agents. In this work, we present a method for predicting realistic motion plans for cooperative human-robot teams on the task. Using a Variational Recurrent Neural Network (VRNN) to model the variation in the trajectory of a human-robot team across time, we are able to capture the distribution over the team’s future states while leveraging information from interaction history. The key to our approach is leveraging human demonstration data to generate trajectories that synergize well with humans during test time in a receding horizon fashion. Comparison between a baseline, sampling-based planner RRT (Rapidly-exploring Random Trees) and the VRNN planner in centralized planning shows that the VRNN generates motion more similar to the distribution of human-human demonstrations than the RRT. Results in a human-in-the-loop user study show that the VRNN planner outperforms decentralized RRT on task-related metrics, and is significantly more likely to be perceived as human than the RRT planner. Finally, we demonstrate the VRNN planner on a real robot paired with a human teleoperating another robot.

\end{abstract}

\section{Introduction}
\par Humans internally develop and rely on models of the world around them to make goal-oriented decisions. They have internalized strategies that drive their decision-making based on their goals and observations. In collaborative tasks, human strategies must accommodate their teammates' behaviors in order to successfully meet their mutual objectives. 
Robots that collaborate with humans must emulate this ability to be effective teammates \cite{Dafoe2020,Kragic2018}. Achieving this capability is especially difficult due to the inherently multimodal nature of human interactions. However, even when robots are equipped with an underlying prior on human actions, human actions are extremely noisy and difficult to predict. This situation is especially true in complex, continuous action and state space tasks. Learning the sampling distributions over future motion sequences from successful human demonstrations not only allows the robot to plan over potential outcomes, but also come up with viable actions that result in more fluent interactions (e.g. not pushing the table towards each other, or pulling in opposite directions).

\begin{figure}[h!]
  \centering
  \includegraphics[width=0.35\textwidth]{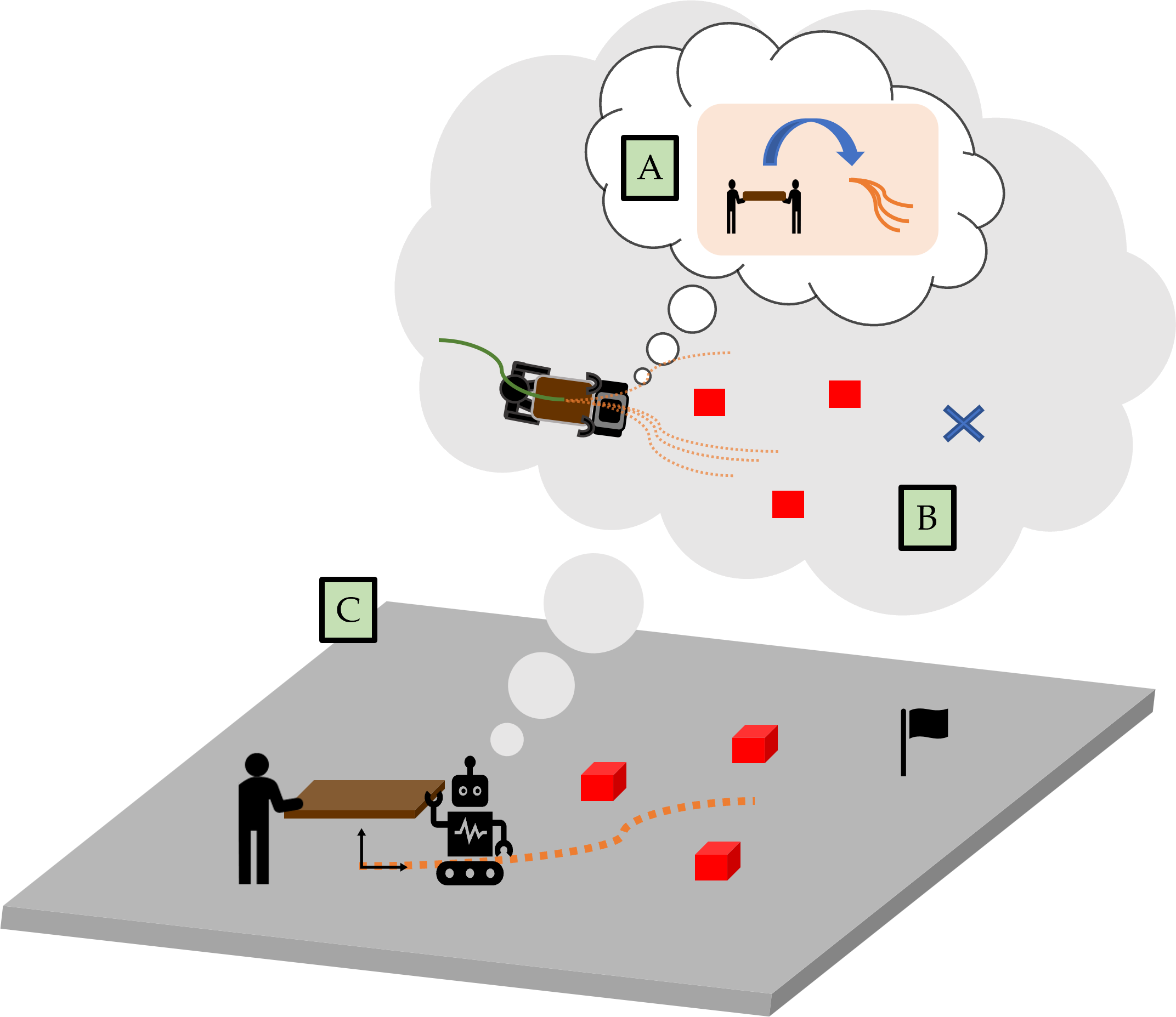}
  \caption{Our method leverages successful human-human demonstrations (solid orange lines) for the cooperative table-carrying task (A). We train a robotic planner that leverages a generative model to sample and reason over paths (dotted orange lines) for cooperative path planning (B). The robot uses the model to perform online receding horizon planning when executing the task with a human in the loop (C).}
  \label{splash}
  \vspace{-0.5cm}
  
\end{figure}

\par An example of a human-robot task that involves both agents exerting joint effort is the human-robot cooperative table-carrying task, which is the primary focus of this paper. As a physically challenging and decentralized collaborative task, both agents must align their goals without explicit verbal communication, and each agent should anticipate and adapt to their collaborator's actions to allow for fluent collaboration. Motivated by this task, we seek to improve online, cooperative planning for robotic assistants by leveraging demonstrations of human-human cooperative behavior to predict realistic future trajectories of human-robot teams navigating the task (Fig. \ref{splash}). The difficulty lies in how each agent's independent action contributes to the joint action, which influences the state of the task due to physical constraints -- the joint state (of both agents) is not a clear signal of how each human individually wants to move. Furthermore, there is a \textit{multimodality} of paths, which are distinct, yet equally valid, that can be taken on this interactive task. 
        
\par \textit{Contributions}: \edits{In this paper, we make two contributions: 1) We provide an open source custom environment for the cooperative table carrying, a continuous state-action task; and 2) We develop a framework for leveraging offline human demonstrations of cooperative behavior by training an sampling-based planner that generates realistic, human-like motion predictions, which samples for re-planning during test time in a receding horizon fashion to successfully cooperate with a human on the task}. 

\par To achieve this, our data-driven framework must capture the following system requirements: (1) variance over time in generated plans, (2) multimodality in interactions, and (3) motion realism to reflect what a human would actually do while cooperating with a partner. Due to the physical and simultaneous nature of the task, our model makes no explicit assumptions about the other human's high-level behaviors, but rather learns from sequences of demonstrated behaviors performed on the task. We leverage this behavior to develop a learned sampling-based planner for a robot agent that learns realistic, human-like motion, as evaluated with several metrics that have not yet, to the best of our knowledge, been deployed to evaluate tasks involving human-robot interaction. Furthermore, we demonstrate that our planner works with a human to successfully complete the table-carrying task in human-in-the-loop scenarios, both in simulation and on real robots.


\begin{figure*}[h!]
\centering
\vspace{0.2cm}

\subfloat[]{\label{fig:network}
\centering
\includegraphics[width=0.7\textwidth]{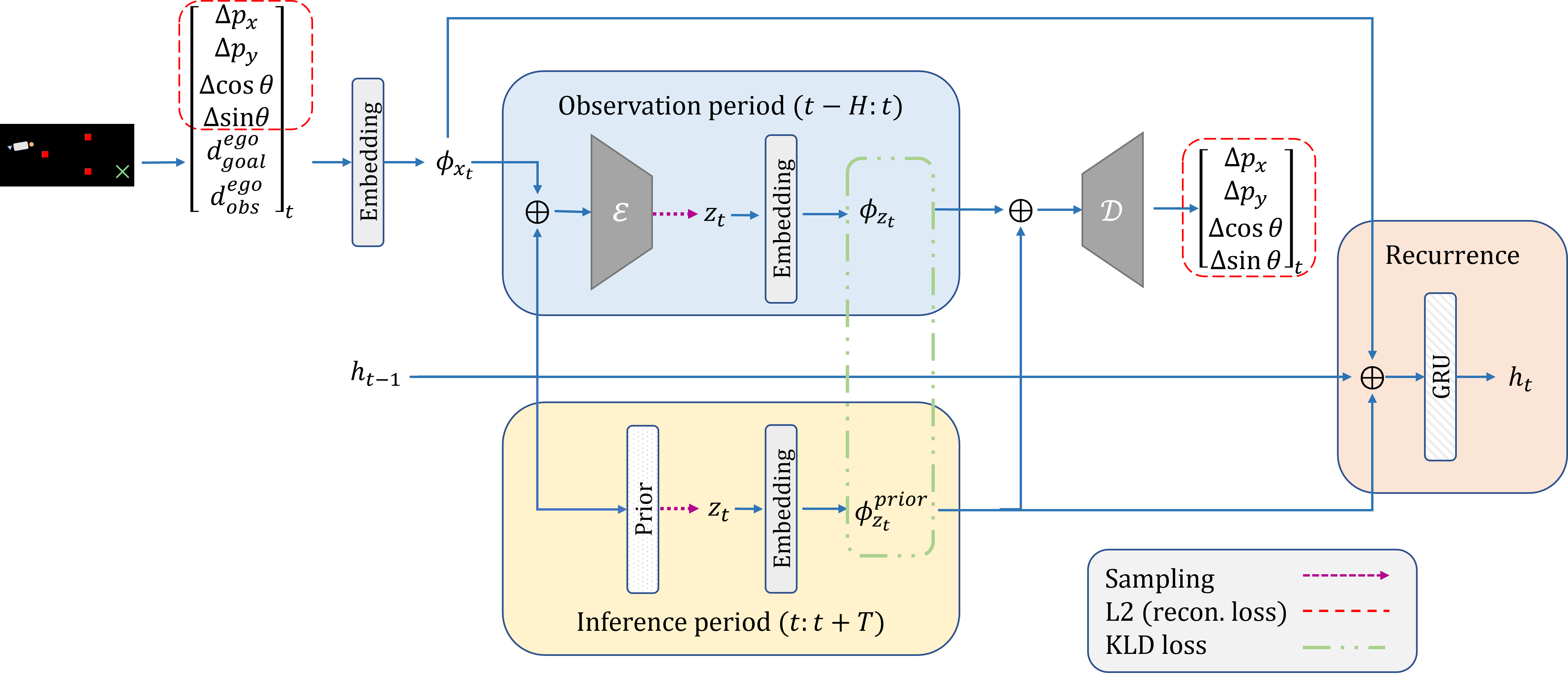}
}
\hfill
\subfloat[]{\label{fig:tabledynamics}
\centering
\vspace{1cm}
\includegraphics[ width=0.22\textwidth]{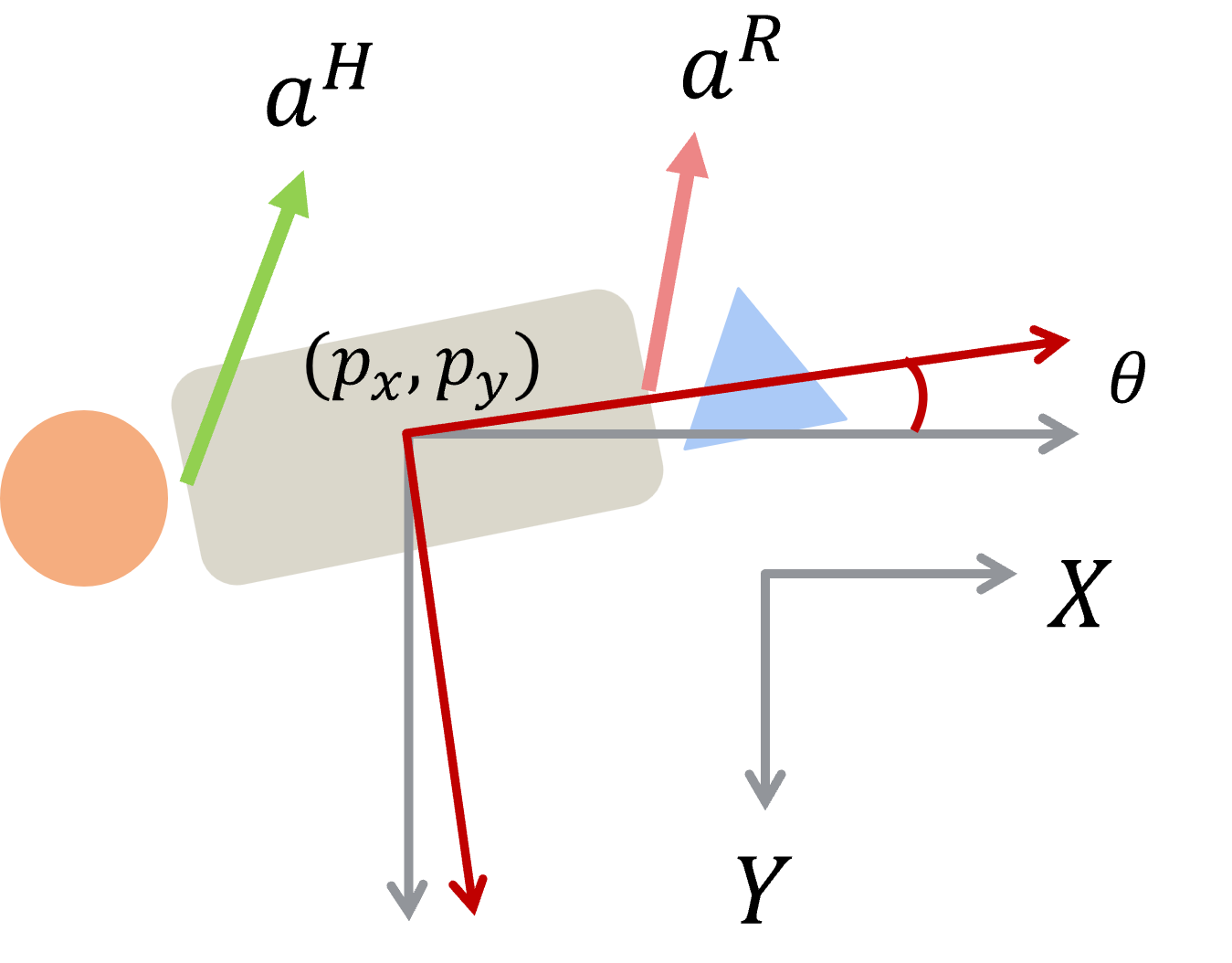}
}
\caption{(a) Model architecture of the VRNN used for cooperative carrying. (b) Table state in simulator. One agent is denoted by an orange circle, the other by a blue triangle.}
\vspace{-0.5cm}

\end{figure*}

\section{Related Work}

\par The problem of modeling and predicting human behavior for interactive tasks is widely studied and has been approached from various viewpoints. For cooperative tasks wherein spatio-temporal relationships with other agents or environments are important considerations, predicting motion is a challenging aspect of the cooperation. 
Early works in human motion prediction use state transition models to encode assumptions of the physical state of the agents. These models have been physics-based, such as the ``social force" model \cite{Helbing_1995}. Temporal networks, such as Recurrent Neural Networks (RNNs), have also been used to capture time dependencies in motion prediction; in the human-robot interaction space, RNNs have been deployed in various trajectory prediction problems \cite{rnn}, \cite{ss-lstm}, \cite{social_lstm}.

\par In working with humans, many techniques have been applied for human intent estimation. Particularly, some works have used Bayesian frameworks \cite{hu2018framework} along with black box modeling techniques \cite{mlp} to capture human intent in human-robot interaction scenarios. Others explicitly break down the task and employ game-theoretic modeling to consider turn-by-turn interactions \cite{fisac}. However, many of these interaction-aware works notably do not capture cooperative behaviors, particularly for tasks such as cooperative table-carrying, which require some degree of modeling the mostly physical and non-verbal \edits{interactions in continuous time}, where decisions occur on the order of $\sim$1 second or less. \edits{In Nikolaidis, et. al. \cite{nikolaidis2017}, the authors use the bounded memory assumption, which addresses the limitations of modeling humans as perfectly rational when making decisions over a limited time span; however, they model the human-robot cooperative carrying task with binary human-robot actions made at discrete time-steps, which does not address real-time fluctuations in decision-making. Several works in physical human-robot interaction (pHRI) for cooperative tasks model low-level actions and ground the task in the sensorimotor space of the robot, and do not capture the multimodality of interactions that could be indicative of potential decision points, such as those created by obstacles in the path \cite{phri-rl}. \cite{bussy} proposes that when a robot guesses its partner's trajectory correctly, then it is able to proactively participate in the task, but assumes that plan to the goal does not change \textit{during} the interaction. Our approach is similar in that the robot learns to predict \textit{team} actions as seen from demonstrations, but we account for live interaction by re-planning via generating new trajectories online. In this way, the notion of leader and follower is abstracted during cooperation, which can be useful when online adaptation is necessary  throughout the interaction. The robot learns from how teams behaved cooperatively and leverages this learned behavior to guide the team towards the joint objective. }

\par Inspired by recent breakthroughs in generative modeling that have shown promising success in modeling the motion prediction problem in interactive scenarios \cite{social_vrnn}, \cite{schmerling2017}, \cite{Zhan2018}, \cite{boris}, \edits{as well as works on learning latent dynamics for planning \cite{hafner2019learning}, \cite{ichter2019robot},} we address the task of cooperative table carrying, wherein two agents navigating a scene cluttered with obstacles must reach their goal destination without collision. Cooperative table-carrying is a challenging sequential task that requires joint effort; therefore, it is important that our model can capture (1) variance over time, (2) multimodality in interactions, and (3) motion realism to reflect what a human would actually do so that the robot can be a better partner.


\section{Problem Formulation: Human-Robot Cooperative Carrying}
\par We consider a single human agent and a single robot agent both maintaining contact at opposite ends of a mechanical load being carried. Unlike in scenarios where it is desirable for each agent to move around without contact with one another~\cite{social_vrnn}, the poses of each agent in the cooperative carrying task are coupled by parameters of the load being carried. The dynamics of the joint state $s_t \in \mathcal{S} \subset \mathbb{R}^n$, which we model as the state of the mechanical load, are 
$$
    \edits{s_{t+1} = f(s_t, a_{t}^R, a_{t}^H) }\eqno{(1)}
$$
where $a_{t}^i \in \mathcal{A}_i \subset \mathbb{R}^{m_i}$ is the action from each agent $i \in \{R, H\}$ at time step $t$.

\par We aim to learn a probabilistic model for the robot that captures the multimodality of future trajectories of the carried load up to time step $T$, $s_{t+1:T}$, conditioned on previous $H$ states of the table, $s_{t-H:t}$, as well as information about the human-robot team's surroundings. We represent knowledge of obstacle and goal locations relative to the table as $d^{ego}_{goal}~\in~\mathbb{R}^2$, which is the heading vector to the goal in the table's ego frame; and $d^{ego}_{obs}\in \mathbb{R}^2$, the heading vector to the nearest (L2 distance) obstacle in the table's ego frame. The task demands accounting for high variability in motion conditioned on environment, as well as multimodality in human actions; for this reason, we choose to include latent hidden variables to model the sequential task, $p(s_{t+1} | h_{t-1}, s_t, z_t)$, where $z_t$ is a latent variable introduced to help structure the learning.

\subsection{Motion Prediction Model}

\par  Considering the temporal nature of the task, we seek to leverage temporal networks for model predictions. Vanilla temporal models such as Recurrent Neural Networks do not possess the capability to model large deviations in observations with a deterministic hidden state; therefore, to capture the variations in the sequence of observations, we adopt the Variational Recurrent Neural Network architecture developed in prior work \edits{ \cite{vrnn}, as seen in Fig. {\ref{fig:network}}}. The model input $s_{t-H:t}$ is an 8-dimensional vector sequence of previous \edits{change in} states over the observational period, $H$:
$$  s_{t-H:t} = 
\begin{bmatrix}
    \Delta p_x &  \Delta p_y & \Delta cos\theta & \Delta sin\theta  & d^{ego}_{goal} & d^{ego}_{obs}
\end{bmatrix}_{t-H:t}^T
$$
where $\Delta p_x$, $\Delta p_y$, $\Delta cos\theta$, $\Delta sin\theta$ ($\theta$ is smoothed via cosine and sine transforms) are the discrete time differences in the world frame of the table's $ x$, $y$, and orientation $\theta$ (Fig. \ref{fig:tabledynamics}), respectively. The model predicts the next change in state, 
$$  s_{t+1:t+T} = 
\begin{bmatrix}
    \Delta p_x &  \Delta p_y & \Delta cos\theta & \Delta sin\theta
\end{bmatrix}_{t+1:t+T}^T
$$

\par \edits{In the training data, trajectory subsequences consist of state history for $H$ timesteps, and are $T$ steps in length total. Here, we denote $\tau$ as the timestep in the subsequence, and $t$ as the \textit{current} timestep from which predictions begin. The forward step of the VRNN is split by time into two parts: first, we pass the previous hidden states, $h_{\tau-1}$, and state history, $s_{\tau-H:t}$, to the encoder for obtaining situational context; then, starting at $\tau = t+1$, we perform rollouts by sampling from only the prior, which conditions on the updated previous hidden states.}

\edits{The network begins with a hidden state initialization of zeros. From $\tau = t-H:t$, inputs to the encoder pass through an embedding layer followed by an encoder, $q(z_{\tau} \mid s_{\leq T}, z_{\leq T})$, and feed into two separate networks that parameterize the Gaussian latent variable $z_\tau$, i.e. the mean and log-variance parameters, each with dimension size 6. From $\tau = t+1:T $, latents are sampled from the prior network, $\phi^{prior}_{z_\tau}(h_{\tau-1})$, which is conditioned on all previous states via the last hidden state, $h_{\tau-1}$. After performing the reparameterization trick, sampled latents pass through an embedding layer and decoder, \mbox{$p(s_\tau \mid z_{\leq\tau}, s_{<\tau})$}, which performs reconstruction. During each $\tau$ in the forward pass, we update the hidden state using a Gated Recurrent Unit (GRU) conditioned on the sampled latent and input used to generate the latent. The encoder and decoder networks are 2-layer MLPs with 128 hidden units; all other networks mentioned are 2-layer MLPs with 64 hidden units.}
\edits{With this framework, we} can roll out future motion plans up to time $T$ by conditioning on observation history. Note that we can vary $H$ and $T$ since we are leveraging a recurrent network structure. The loss function is given by minimizing a  variational lower-bound per timestep $\tau$:

\vspace{-0.25cm}

\begin{equation}
\begin{split}
    \mathbb{E}_{q(z_{\leq T} \mid s_{\leq T})}  = & \sum_{\tau=1}^{T} \Bigl( \log p(s_\tau \mid z_{\leq\tau}, s_{<\tau})  \\
    &  - KL( q(z_{\tau} \mid s_{\leq \tau}, z_{< \tau}) \big\| p(z_\tau \mid s_{\leq\tau}, z_{<\tau})) \Bigr)
\end{split}
\vspace{-2mm}
\end{equation}
\vspace{-2mm}

where the log term becomes the L2 loss on the predicted states.

\begin{figure}[h!]
\centering
\subfloat[RRT plans.]{\label{fig:capparatus}
\centering
\includegraphics[width=0.45\linewidth]{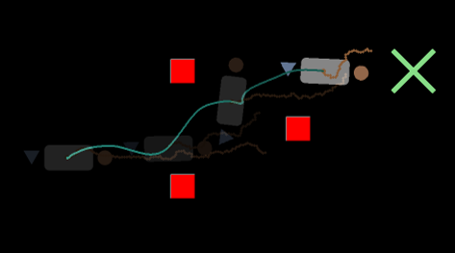}
}
\hfill
\subfloat[VRNN plans.]{\label{fig:cdiagram}
\centering
\includegraphics[width=0.45\linewidth]{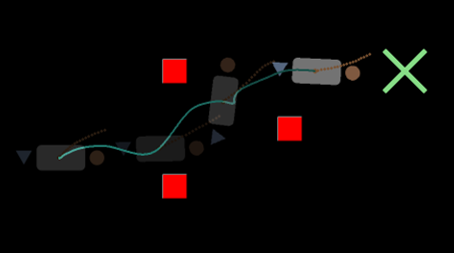}
}
\caption{Table state predictions (orange lines) plotted along a previously collected demonstration trajectory (ground truth, green line) at various time steps on a map unseen during model training. The red squares denote the obstacles, the green cross denotes the goal.}
\vspace{-0.5cm}

\label{fig:pred_traj}
\end{figure}

\subsection{Training and Validation of Prediction Model}

\par  We implemented a \edits{\href{https://github.com/eleyng/table-carrying-ai}{custom gym environment} for the cooperative table-carrying task, which allows for data collection and planner evaluation. Akin to the Overcooked AI environment in \cite{carroll2020utility}, this environment can serve as a benchmark environment for challenging coordination problems. This environment has a \textit{continuous state-action space} for \textit{joint-action} tasks.} Given input forces from two agents, \edits{$a^{H}$ and $a^{R}$}, we model the table to follow double-integrator dynamics, where \edits{$a^i, i \in{R, H}$} are inertial forces that contribute to table acceleration. We add damping parameters on table velocity to mimic frictional forces. The simulator records force inputs and the table state at a maximum frequency of 30Hz. Actions are recorded from both humans during dataset collection using joystick controllers. The action space is continuous, with \edits{$a \in [-1, 1]$} normalized with the joystick maximum and minimum displacement in both the $x$ and $y$ axes.

\par To train our model, we collected a human-human demonstration dataset from the simulator on a variety of initial and goal conditions, and maps. At the start of each demonstration, we initialize the simulator randomly, parameterized by 3 different initial table poses, 7 different obstacle configurations, and 3 different goal poses -- thus, a total of 63 different possible configurations. The maps are configured such that the agents always generally start at the left side of the map, and navigate through an obstacle-filled space to a goal location on the right side of the map. A total of \edits{339} trajectories of human demonstrations on 5 different pairs of people were collected \edits{for training and validation, with a 80:20 split. For planner evaluation (predictions only, without human-in-the-loop), 30 human-human demonstrations were collected as ground truth trajectories: }15 of them\edits{, which included maps layouts seen during training and validation,} were separated into a \textit{holdout test set}; another 15 trajectories were demonstrations on \textit{unseen maps} with novel initial position, obstacle layouts, and goal configurations to which the model did not have exposure \edits{during training and validation. Notably, in the unseen maps, obstacles were placed on routes which were often taken by human-human pairs in the demonstrations on the training maps.} For each map, we told the demonstrators to non-verbally and cooperatively move the table together from the start to end location while avoiding obstacles. The models were trained on a 12th Gen. Intel i9 12900K processor with 2 NVIDIA GeForce 2080 Ti GPUs.


\section{Experiments}
\subsection{Experimental Setup}

\par We evaluated the model from three aspects: (1) quality of trajectory predictions generated, (2) ability to navigate human-in-the-loop environments, and (3) quality of interaction. To evaluate the model's trajectory prediction, we compare sampled rollouts against 15 human-human demonstrations on unseen, novel maps. \edits{To evaluate the quality of the model as a planner, we also conduct a user study with human-in-the-loop experiments to study trajectories from the human-robot team. We also conduct a Turing Test to evaluate the perceived human-likeness of the planner.}

\par \edits{As a baseline, we use a sampling-based algorithm, Rapidly-exploring Random Tree (RRT), to sample paths for both agents to follow. We compare the motion realism and diversity of generated rollouts from our method with those generated by an ``ideal" team that leverages centralized planning.  During the human-in-the-loop evaluations, we use RRT in a decentralized fashion as a baseline (referred to as Dec-RRT): one agent plans with RRT while the other is teleoperated by a human through a joystick.}

\par
\begin{figure}[h!]
  \centering
  \includegraphics[width=0.45\textwidth]{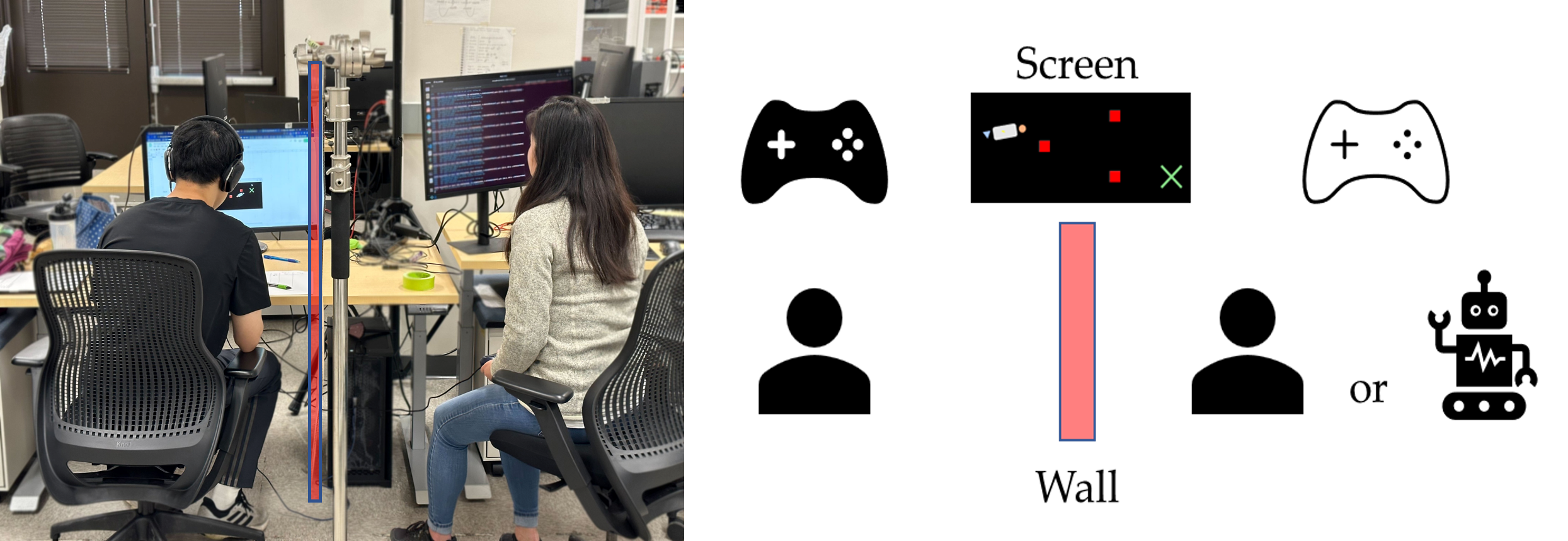}
  \caption{Human-in-the-loop experimental setup, in reality (left) and in concept (right). For the Turing test, we purposefully block audio-visual observations of the partner from the human subject using headphones and a wall (shaded in red) separating both parties.}
  \vspace{-0.25cm}
  \label{turing_setup}
\end{figure}

 \edits{For the user study, 15 participants completed the cooperative carrying task over 60 trials. Two participants were involved in the human-human demonstrations. The experimental setup is shown in Fig. \ref{turing_setup}, wherein the participant is unaware of who their partner is. 15 of the trials were with the Dec-RRT planner, and another 15 trials were with the VRNN planner, all on the same unseen map configurations for each subject. This was repeated on a set of test holdout maps. Of each 15 trial segment, 5 were selected to play with a real human partner, who was the same person across subjects. At the end of each trial, the participant was asked whether they thought they were playing with a human or a robot.}

\begin{figure*}[!h]
  \centering
  \vspace{2mm}
  \adjincludegraphics[width=0.95\textwidth, trim={{0.0\width} {0.1\height} {0.0\width} {0.1\height}}, clip]{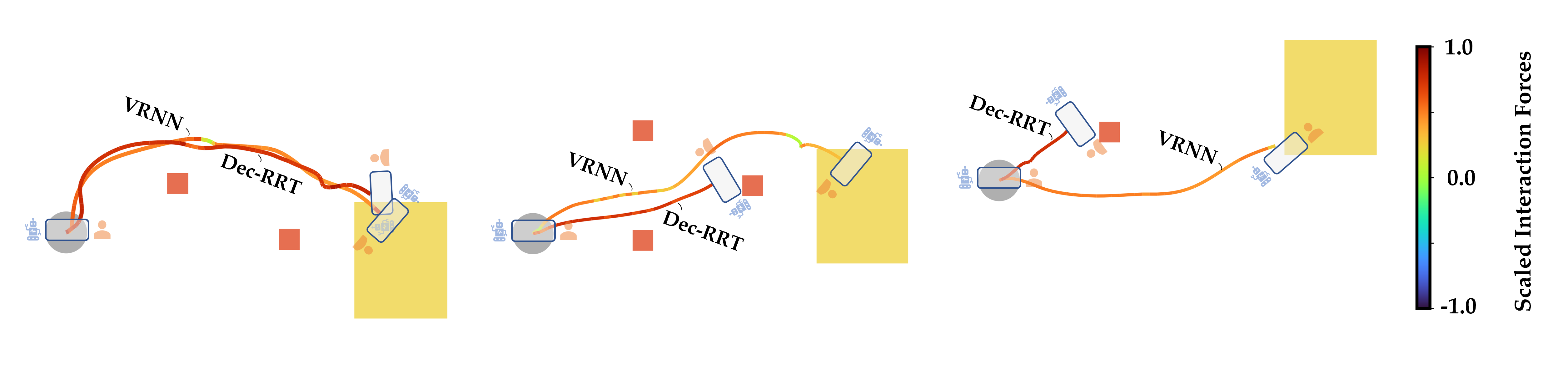}
  \vspace{-2mm}
  \caption{Interaction forces between planner and real human across a selection of human-in-the-loop trials, scaled by min-max interaction forces over all trials. Non-zero interaction forces indicate dissent. (Left) In one test holdout map, the team chooses the same path above the obstacle, but the VRNN planner clearly contributes lower interaction forces over the trajectory. (Middle) In one unseen map configuration, the VRNN rotates with the human partner to avoid the obstacle, while the Dec-RRT planner does not create enough space to avoid it. (Right) In another unseen map configuration, the Dec-RRT planner chooses to move above the obstacle, but fails to negotiate with the human in time to avoid it.}
  \label{fig:inter_f}
  \vspace{-2mm}
\end{figure*}

\par
\begin{figure}[h!]
  \centering
  \includegraphics[width=0.35\textwidth]{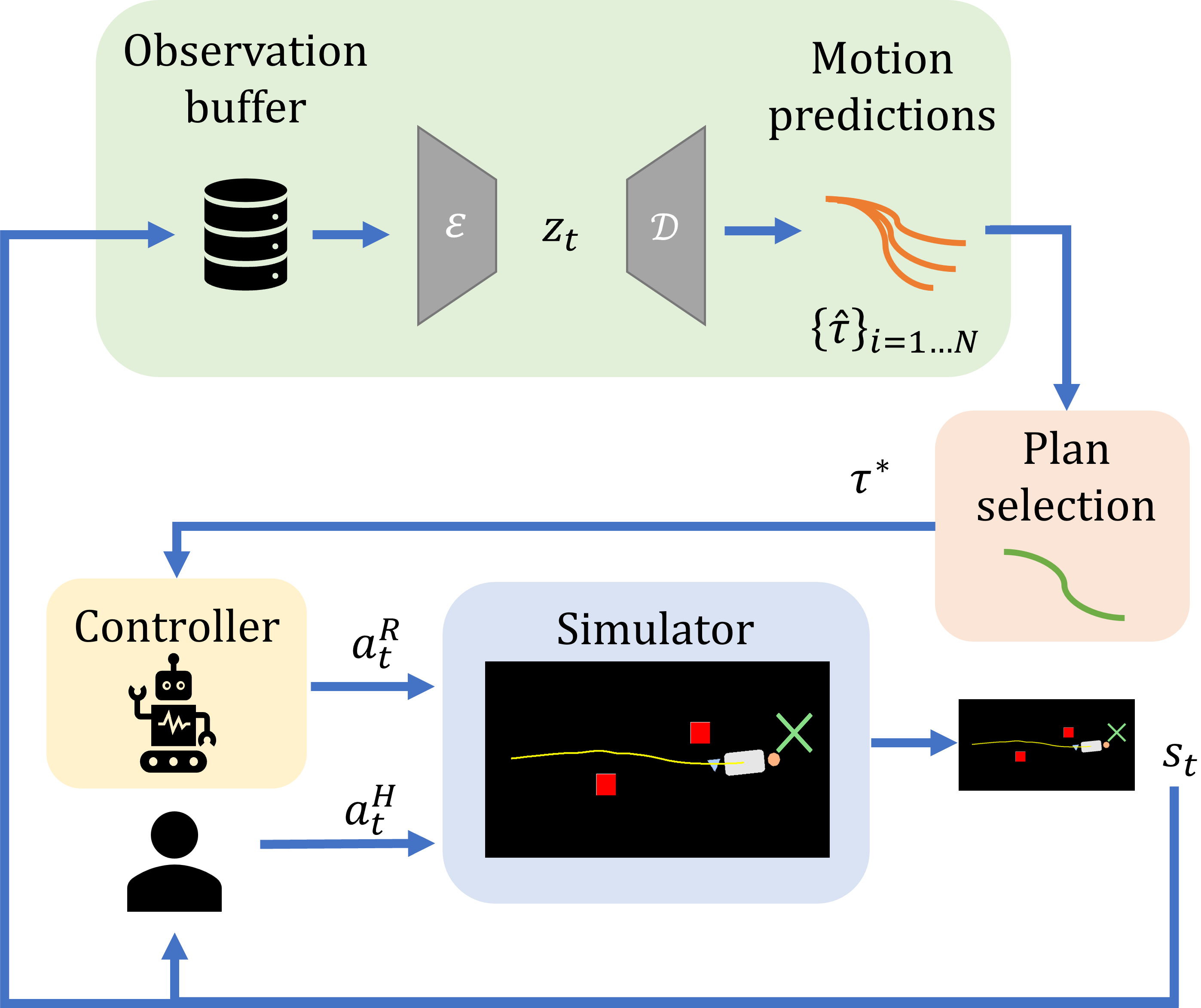}
  \caption{Our framework leveraging learned motion prediction models for planning and control.}
  \vspace{-0.5cm}
  \label{fig:framework}
\end{figure}

 Fig. \ref{fig:framework} shows the human-in-the-loop framework when the planners are operating. During prediction with the VRNN planner, we provide the model with the first second (30 steps in the simulator with 30 FPS) of the ground truth trajectory. Then, \edits{starting from the current time $t$, we} sample rollouts for up to $3$ seconds. \edits{We store a running queue of the past 30 simulator steps for the model.}   The agent re-plans \edits{in receding horizon fashion by sampling a batch of trajectory predictions, selecting the trajectory leading to the minimum cost as defined by the environment rewards, and using a simple proportional controller to navigate to the first waypoint predicted by the planner from the time $t$ the plan was computed.  With this framework, future work can optimize over other user-defined rewards relating to the quality of cooperation, such as interaction metrics \cite{hoffman2019evaluating}}.

\begin{table*}[h!]
\centering
\vspace{3.5mm}
\begin{tabular}{|| l | c c c c c | c c c c c ||} 
\hline
& L2 & FD $(10^3)$ & Var $x$ & Var $y$ & Var $\theta$  \\
\hline
Centralized RRT & 952.41 & 1325.63 & 295.88 & 37.81 & 1.47 \\
VRNN & 751.02 & 892.19 & 991.01 & 97.74 & 1.59 \\
\hline
\end{tabular}
\caption{Comparison of plans generated from the VRNN and centralized RRT planners, with playback ground truth data from human-human demonstrations (no human in the loop).}
\label{table:plans}
\end{table*}

\begin{table*}[h!]
\centering
\begin{tabular}{|| l | c | c c | c c | c c ||} 
 \hline
  & \multicolumn{1}{c|}{ } & \multicolumn{2}{c|}{$x$} & \multicolumn{2}{c|}{$y$} & \multicolumn{2}{c||}{$\theta$} \\
& L2 & FD $(10^3)$   & Var  & FD $(10^3)$  & Var & FD & Var  \\ 
 \hline
 GT & -- & -- & 1218.11 & -- & 221.55  &  --  &  1.92\\ 
 Dec-RRT & 343.02 & 243.13 & 758.00 & 65.46 & 86.33  & 1.75  & 0.56  \\ 
VRNN & 388.78 & 106.40 & 1100.27 & 113.93 & 99.78  & 0.73  & 1.53  \\  [1ex] 
 \hline
\end{tabular}
\caption{Comparison of trajectories collected from human-in-the-loop interactions on unseen map configurations, wherein the human plays with either the VRNN planner or the baseline Decentralized RRT (Dec-RRT) planner.}
\vspace{-2mm}
\label{table:hil}
\end{table*}
  
\subsection{Evaluation Metrics}

\par Quantifying rollouts and interactions of a human-robot interactive task is a difficult problem that requires evaluation on multiple fronts. Our metrics are based on prior work in motion realism from computer vision literature, as well as task-based metrics such as success rate and time to completion\edits{, and qualitative metrics via a Turing Test for joint-action tasks}. We consider ideal cooperative behavior to be (1) human-like, or as close as possible to what two humans cooperating on the task would do in a demonstration; (2) diverse, such that a variety of behaviors can emerge; and (3) coordinating, such that it is able to successfully complete the task together.

\par We evaluate the plans generated by our model on a ground truth (GT) test dataset of demonstrations on challenging, unseen map configurations, as well as on real-time human interactions using the following metrics in SE(2) space ($p_x, p_y, \theta$):

\begin{itemize}
  \item \textit{L2}: Distance to ground truth motion\edits{, measured at each time step and summed over each trajectory. The average over the batch of trajectories is reported.}
  \item \textit{Fr\'echet distance (FD)}: FD can be used to capture motion realism by measuring the distribution distance between generated and ground-truth motion sequences as used in \cite{learntodance}, \cite{ng}. We calculate FD between the distributions of each planner's rollouts and trajectories in the test set. For human-in-the-loop trials, we further capture the distribution difference on each dimension.
  \item \textit{Variance}: Each planner's ability to generate diversity in motion is characterized by computing the batch-averaged temporal variance on each dimension.
\end{itemize}


For the human-in-the-loop trials, we also use task metrics to analyze each planner's ability to coordinate with a human:

\begin{itemize}
    \item \textit{Success rate (\%)}: The percentage of runs in which the human and planner successfully reached the goal location.
    \item \textit{Time to completion (s)}: The time in seconds it took to successfully complete the task. 
\end{itemize}

\edits{Finally, after each trial, we ask participants to respond to the question, "Do you think that the other player was controlled by a human or a robot?" }

\subsection{Comparison of Sampled Plans}

\par Table \ref{table:plans} shows a comparison between plans generated by the VRNN and plans generated by the baseline Centralized RRT. The high FD score on the centralized baseline RRT suggests that while RRT can plan for task success, the distribution of plans generated by the RRT planner is further from the distribution of human-human demonstrations, signaling that centralized behavior from high-performing motion planners is not representative of ground truth cooperative behavior in human-human demonstrations. Fig. \ref{fig:pred_traj} visually shows the differences in plans generated from each planner.

\subsection{User Study: Human-in-the-loop Trials}

\par Evaluation of the VRNN planner in simulation with a human-in-the-loop suggests that it is capable of generating more human-like motion plans than the baseline planner, as indicated by lower FD scores in all but the y dimension. The VRNN planner also exhibits higher diversity in pose than the Dec-RRT planner, and outperforms the Dec-RRT planner on both task success rate and time taken to complete the task \edits{(Table \ref{table:taskmetrics})}. The VRNN planner's better performance on task metrics suggests that while trajectories resulting from its cooperation with a human partner were further from the ground truth human-human demonstrations as measured by the L2 distance, its capability to exhibit higher variance across $x$, $y$, and rotation $\theta$ contributes to its higher success rate on unseen map configurations.

\par
\begin{figure}[h!]
  \centering
  \vspace{1mm}
  \includegraphics[width=0.45\textwidth]{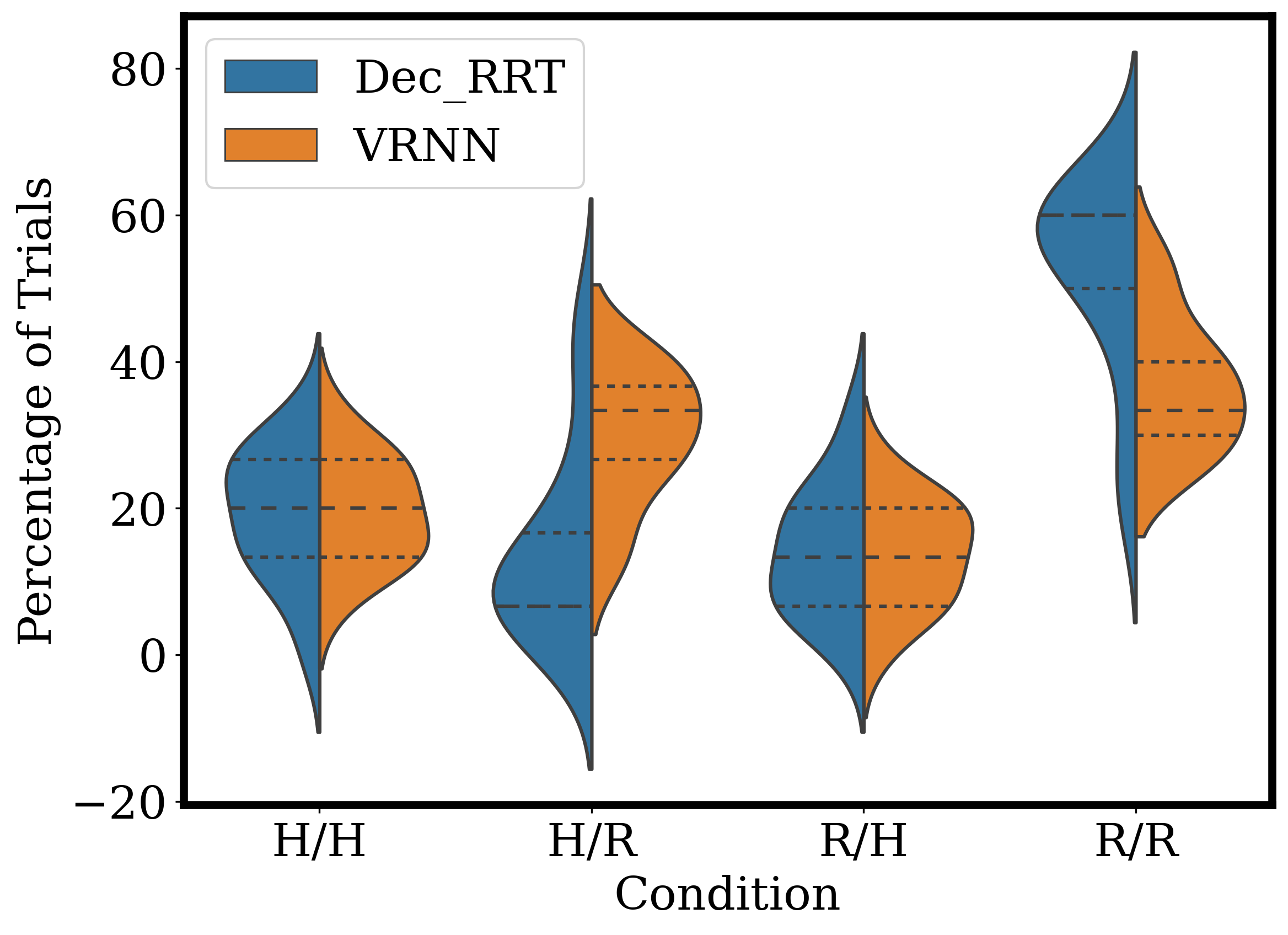}
  \caption{Violin plots depicting average and interquartile ranges of predictions of partner type versus actual partner type as percentages of 15 test trials per subject ($N=15$) for the Turing Test classification on unseen maps, i.e. H/R refers to predicted human partner vs. actual robot partner, etc. The conditions sum to 100\% per planner type per subject.}
  \vspace{-0.25cm}
  \label{fig:violin}
\end{figure}

 \edits{To gauge whether the VRNN planner was perceived as more "human-like" than the Dec-RRT planner, we ran a two-sample t-test on the participants' responses to the question of whether they thought they were playing with a human or a robot. Fig. \ref{fig:violin} depicts the classification performance statistics across planner type and subjects. Results show that there is a significant difference in how the two planners were perceived, with the VRNN partner more likely to be mistaken to be human than the Dec-RRT planner ($45\%$ vs. $22\%$, $p < .001, d = 0.2267, N = 15$).  As to why this may have been the case, we examined the interaction forces on the table between the human and robot. Interaction forces, as computed in \cite{vijay}, do not contribute to motion, and non-zero values result in compression or stretch of the load being carried. We scaled the interaction forces by the peak min and max interaction forces across all trials, and visualized them along several trajectories in Fig. \ref{fig:inter_f}. The VRNN planner clearly exhibited interaction forces closer to zero over longer segments of the trajectories than Dec-RRT, suggesting that it has better coordination capability, through timing and precision of actions.}

\begin{table}[h!]
\centering
\resizebox{1.\columnwidth}{!}{%
\begin{tabular}{|| l | c c | c c ||} 
\hline
& \multicolumn{2}{c|}{Holdout set} & \multicolumn{2}{c||}{Unseen map} \\
& Success ($\%$) & Time (s) & Success ($\%$)  & Time (s) \\
\hline
Dec-RRT & 68.00 & 18.71 $\pm$ 3.38 &  24.00 & 20.59 $\pm$ 4.86 \\
VRNN & 76.00 & 17.27 $\pm$ 4.29 & 62.67 & 16.29 $\pm$ 3.66 \\
\hline
\end{tabular}
}

\caption{Task success rate ($\%$) and average time (s) to completion for successful trajectories, i.e. the human-robot team reached the goal without colliding into obstacles.}
\label{table:taskmetrics}

\vspace{-4mm}
\end{table}

\begin{figure}[!h]
  \centering
  \vspace{1mm}
  \adjincludegraphics[width=\columnwidth, trim={{0.0\width} {0.0\height} {0.0\width} {0.0\height}}, clip]{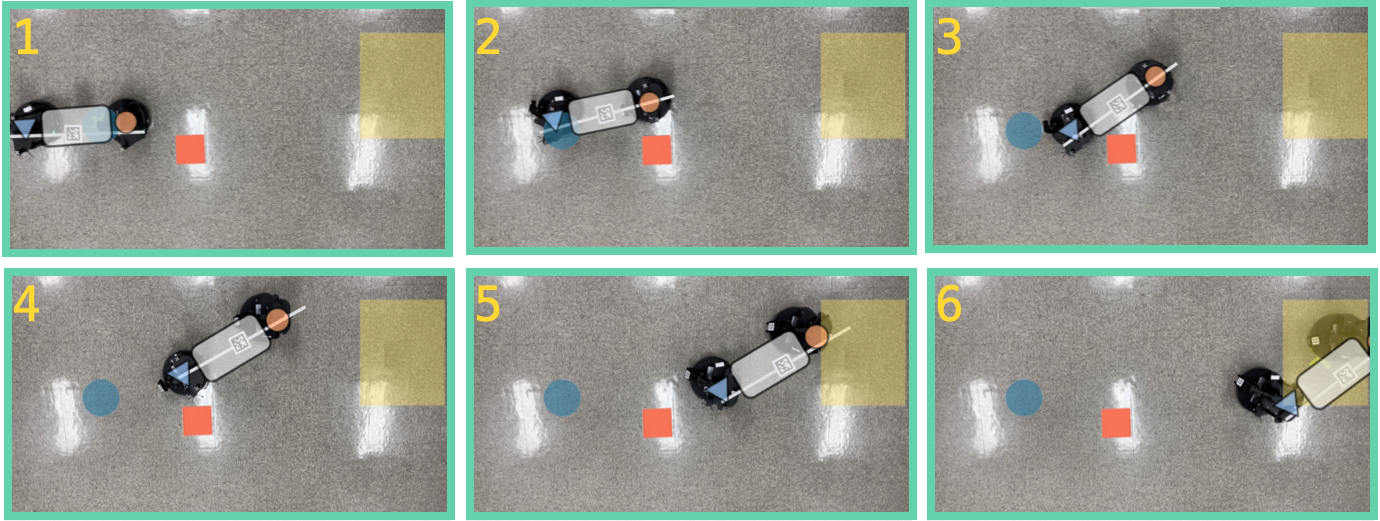}
  \caption{Sim-to-real transfer of the VRNN planner trained in simulation to control a LoCoBot in the real world. The VRNN planner (blue \edits{triangle}) successfully cooperates with the human subject (orange circle) in carrying the table from the initial position (blue circle) to the goal region (yellow rectangle), while avoiding the obstacle (red rectangle).}
  \label{fig:real_demo}
\vspace{-0.55cm}
\end{figure}

\subsection{Real robot demonstration}

\par We demonstrate the utility of our method on a real robot setup. Using two Interbotix LoCoBots, we show that our framework can be used to predict motion and adapt to real-time human inputs shown in Fig. \ref{fig:real_demo}. The human provides joystick inputs while the robot leverages the planner to anticipate the team's motion. The robots successfully navigate an unseen map scenario. For video of real robot results, please see the \edits{\href{https://youtu.be/CqWh-yWOgeA}{supplemental video}}.

\section{Conclusion}

\par In this work, we developed an evaluated a planner for human-robot cooperative load-carrying, which leverages human demonstration data to predict realistic motion plans. We demonstrated in two different evaluations (i.e. generated plans and trajectories from human-in-the-loop tests) that paths generated by the VRNN model are similar to the human demonstration data, have large variance in motion, and can synthesize well with humans on novel maps. We also demonstrate a successful zero-shot transfer of the model to a real robot navigating an unseen map with a human teleoperating another robot. Furthermore, we provide an open source custom gym environment for this task.

\par Future work will improve the observation model used for this task. Furthermore, we did not leverage action information. There may be more information on intent that can be gathered from observing actions, despite being highly noisy. Future work will involve leveraging more powerful generative models to learn to predict future action trajectories, and using it for online receding horizon planning in cooperative human-robot tasks.

\addtolength{\textheight}{-4cm}   


\bibliographystyle{unsrt}
\bibliography{refs}

\begin{thebibliography}{10}

\bibitem{Dafoe2020}
Allan Dafoe, Edward Hughes, Yoram Bachrach, Tantum Collins, Kevin~R McKee,
  Joel~Z Leibo, Kate Larson, and Thore Graepel.
\newblock Open problems in cooperative ai.
\newblock {\em arXiv preprint arXiv:2012.08630}, 2020.

\bibitem{Kragic2018}
Danica Kragic, Joakim Gustafson, Hakan Karaoguz, Patric Jensfelt, and Robert
  Krug.
\newblock {Interactive, Collaborative Robots: Challenges and Opportunities}.
\newblock {\em International Joint Conference on Artificial Intelligence},
  2018-July:18--25, 2018.

\bibitem{Helbing_1995}
Dirk Helbing and P{\'{e}}ter Moln{\'{a}}r.
\newblock Social force model for pedestrian dynamics.
\newblock {\em Physical Review E}, 51(5):4282--4286, May 1995.

\bibitem{rnn}
Stefan Becker, Ronny Hug, Wolfgang H{\"u}bner, and Michael Arens.
\newblock An evaluation of trajectory prediction approaches and notes on the
  trajnet benchmark.
\newblock {\em arXiv preprint arXiv:1805.07663}, 2018.

\bibitem{ss-lstm}
Hao Xue, Du~Q Huynh, and Mark Reynolds.
\newblock Ss-lstm: A hierarchical lstm model for pedestrian trajectory
  prediction.
\newblock In {\em 2018 IEEE Winter Conference on Applications of Computer
  Vision (WACV)}, pages 1186--1194. IEEE, 2018.

\bibitem{social_lstm}
Alexandre Alahi, Kratarth Goel, Vignesh Ramanathan, Alexandre Robicquet,
  Li~Fei-Fei, and Silvio Savarese.
\newblock Social lstm: Human trajectory prediction in crowded spaces.
\newblock In {\em 2016 IEEE Conference on Computer Vision and Pattern
  Recognition (CVPR)}, pages 961--971, 2016.

\bibitem{hu2018framework}
Yeping Hu, Wei Zhan, and Masayoshi Tomizuka.
\newblock A framework for probabilistic generic traffic scene prediction.
\newblock In {\em 2018 21st International Conference on Intelligent
  Transportation Systems (ITSC)}, pages 2790--2796. IEEE, 2018.

\bibitem{mlp}
Seungje Yoon and Dongsuk Kum.
\newblock The multilayer perceptron approach to lateral motion prediction of
  surrounding vehicles for autonomous vehicles.
\newblock In {\em 2016 IEEE Intelligent Vehicles Symposium (IV)}, pages
  1307--1312, 2016.

\bibitem{fisac}
Jaime~F. Fisac, Eli Bronstein, Elis Stefansson, Dorsa Sadigh, S.~Shankar
  Sastry, and Anca~D. Dragan.
\newblock Hierarchical game-theoretic planning for autonomous vehicles.
\newblock In {\em 2019 International Conference on Robotics and Automation
  (ICRA)}, pages 9590--9596, 2019.

\bibitem{nikolaidis2017}
Stefanos Nikolaidis, David Hsu, and Siddhartha Srinivasa.
\newblock {Human-robot mutual adaptation in collaborative tasks: Models and
  experiments}.
\newblock {\em The International Journal of Robotics Research},
  36(5-7):618--634, 2017.

\bibitem{phri-rl}
Ali Ghadirzadeh, Judith B{\"u}tepage, Atsuto Maki, Danica Kragic, and
  M{\aa}rten Bj{\"o}rkman.
\newblock A sensorimotor reinforcement learning framework for physical
  human-robot interaction.
\newblock In {\em 2016 IEEE/RSJ International Conference on Intelligent Robots
  and Systems (IROS)}, pages 2682--2688. IEEE, 2016.

\bibitem{bussy}
Antoine Bussy, Pierre Gergondet, Abderrahmane Kheddar, François Keith, and
  André Crosnier.
\newblock Proactive behavior of a humanoid robot in a haptic transportation
  task with a human partner.
\newblock In {\em 2012 IEEE RO-MAN: The 21st IEEE International Symposium on
  Robot and Human Interactive Communication}, pages 962--967, 2012.

\bibitem{social_vrnn}
Bruno~Ferreira de~Brito, Hai Zhu, Wei Pan, and Javier Alonso-Mora.
\newblock Social-vrnn: One-shot multi-modal trajectory prediction for
  interacting pedestrians.
\newblock In {\em Conference on Robot Learning}, pages 862--872. PMLR, 2021.

\bibitem{schmerling2017}
Edward Schmerling, Karen Leung, Wolf Vollprecht, and Marco Pavone.
\newblock Multimodal probabilistic model-based planning for human-robot
  interaction.
\newblock In {\em 2018 IEEE International Conference on Robotics and Automation
  (ICRA)}, pages 3399--3406. IEEE, 2018.

\bibitem{Zhan2018}
Eric Zhan, Stephan Zheng, Yisong Yue, Long Sha, and Patrick Lucey.
\newblock {Generative Multi-Agent Behavioral Cloning}.
\newblock {\em arXiv}, May 2018.

\bibitem{boris}
Boris Ivanovic and Marco Pavone.
\newblock The trajectron: Probabilistic multi-agent trajectory modeling with
  dynamic spatiotemporal graphs.
\newblock In {\em Proceedings of the IEEE/CVF International Conference on
  Computer Vision}, pages 2375--2384, 2019.

\bibitem{hafner2019learning}
Danijar Hafner, Timothy Lillicrap, Ian Fischer, Ruben Villegas, David Ha,
  Honglak Lee, and James Davidson.
\newblock Learning latent dynamics for planning from pixels.
\newblock In {\em International conference on machine learning}, pages
  2555--2565. PMLR, 2019.

\bibitem{ichter2019robot}
Brian Ichter and Marco Pavone.
\newblock Robot motion planning in learned latent spaces.
\newblock {\em IEEE Robotics and Automation Letters}, 4(3):2407--2414, 2019.

\bibitem{vrnn}
Junyoung Chung, Kyle Kastner, Laurent Dinh, Kratarth Goel, Aaron~C Courville,
  and Yoshua Bengio.
\newblock A recurrent latent variable model for sequential data.
\newblock {\em Advances in neural information processing systems}, 28, 2015.

\bibitem{carroll2020utility}
Micah Carroll, Rohin Shah, Mark~K. Ho, Thomas~L. Griffiths, Sanjit~A. Seshia,
  Pieter Abbeel, and Anca Dragan.
\newblock On the utility of learning about humans for human-ai coordination,
  2020.

\bibitem{hoffman2019evaluating}
Guy Hoffman.
\newblock Evaluating fluency in human--robot collaboration.
\newblock {\em IEEE Transactions on Human-Machine Systems}, 49(3):209--218,
  2019.

\bibitem{learntodance}
Ruilong Li, Shan Yang, David~A Ross, and Angjoo Kanazawa.
\newblock Ai choreographer: Music conditioned 3d dance generation with aist++.
\newblock In {\em Proceedings of the IEEE/CVF International Conference on
  Computer Vision (ICCV)}, pages 13401--13412, 2021.

\bibitem{ng}
Evonne Ng, Hanbyul Joo, Liwen Hu, Hao Li, Trevor Darrell, Angjoo Kanazawa, and
  Shiry Ginosar.
\newblock Learning to listen: Modeling non-deterministic dyadic facial motion.
\newblock In {\em Proceedings of the IEEE/CVF Conference on Computer Vision and
  Pattern Recognition (CVPR)}, pages 20395--20405, June 2022.

\bibitem{vijay}
V.R. Kumar and K.J. Waldron.
\newblock Force distribution in closed kinematic chains.
\newblock {\em IEEE Journal on Robotics and Automation}, 4(6):657--664, 1988.

\end{thebibliography}

\end{document}